\documentclass[letterpaper, 10 pt, conference]{ieeeconf}  

\IEEEoverridecommandlockouts                              

\usepackage{graphicx}
\usepackage{color}
\usepackage{caption}

\title{\LARGE \bf
Towards Automated Chicken Deboning via Learning-based Dynamically-Adaptive 6-DoF Multi-Material Cutting
}

\author{Zhaodong Yang$^1$, Ai-Ping Hu$^{1,2}$, Harish Ravichandar$^1$ \\
$^1$Georgia Institute of Technology, $^2$Georgia Tech Research Institute \\
    {\small \texttt{\{halyang, ahu6,  harish.ravichandar\}@gatech.edu}} \\
}

\begin{document}

\maketitle
\thispagestyle{empty}
\pagestyle{empty}

\begin{abstract}

Automating chicken shoulder deboning requires precise 6-DoF cutting through a partially occluded, deformable, multi-material joint, since contact with the bones presents serious health and safety risks. 
Our work makes both systems-level and algorithmic contributions to train and deploy a reactive force-feedback cutting policy that dynamically adapts a nominal trajectory and enables full 6-DoF knife control to traverse the narrow joint gap while avoiding contact with the bones. 
First, we introduce an open-source custom-built simulator for multi-material cutting that models coupling, fracture, and cutting forces, and supports reinforcement learning, enabling efficient training and rapid prototyping.
Second, we design a reusable physical testbed to emulate the chicken shoulder: two rigid ``bone” spheres with controllable pose embedded in a softer block, enabling rigorous and repeatable evaluation while preserving essential multi-material characteristics of the target problem. 
Third, we train and deploy a residual RL policy, with discretized force observations and domain randomization, enabling robust zero-shot sim-to-real transfer and the first demonstration of a learned policy that debones a real chicken shoulder. 
Our experiments in our simulator, on our physical testbed, and on real chicken shoulders show that our learned policy reliably navigates the joint gap and reduces undesired bone/cartilage contact, resulting in up to a 4x improvement over existing open-loop cutting baselines in terms of success rate and bone avoidance. Our results also illustrate the necessity of force feedback for safe and effective multi-material cutting. The project website is at https://hal-zhaodong-yang.github.io/MultiMaterialWebsite/.

\end{abstract}

\section{INTRODUCTION}

\begin{figure*}
    \centering
    \includegraphics[width=0.9\textwidth]{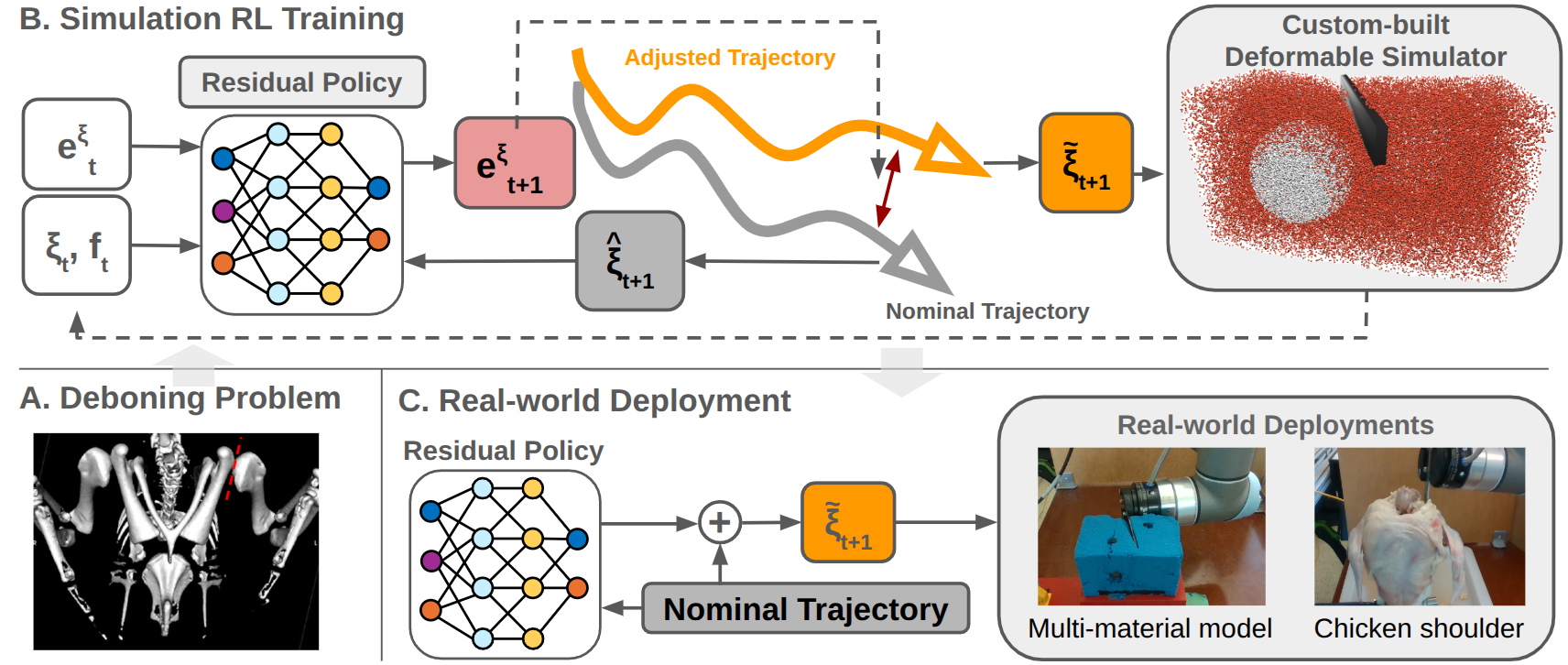}
    \caption{Overview of this paper. A. MRI scan of a chicken. The red line indicates the shoulder joint position and orientation, where we need to precisely cut through. B. We first train the residual policy in the simulator. For each time step, the residual policy observes the current knife pose $\xi_t$ and knife force sensor reading $f_t$ from the simulator. It also observes residual pose $e^\xi_{t}$ from the previous time step, and one step future knife pose $\hat{\xi}_{t+1}$ retrieved from the nominal trajectory. The residual policy eventually produces the next residual pose: $e^\xi_{t+1}$. And $\tilde{\xi}_{t+1} = \hat{\xi}_{t+1} + e^\xi_{t+1}$ is the adjusted pose of the nominal trajectory, also the next knife target pose. We use different nominal trajectories for different cutting tasks. Nominal trajectory is an interpolated trajectory of several keyframe poses, which are either fixed or mapped from the bone position. C. We deploy the residual policy on two challenging real-world multi-material cutting tasks.}
    \label{fig:overview}
    \vspace{-3mm}
\end{figure*}

\hspace{-\parindent}Many tasks in food processing, healthcare, and manufacturing require separating or removing specific parts from objects made of multiple materials.
Chicken deboning is a particularly impactful and challenging example, in which a knife must precisely cut through the shoulder joint of the bird (see Fig.~\ref{fig:overview}A) to separate the chicken's wing from its body without cutting into the shoulder bones or cartilage that are concealed beneath soft tissue and skin. A failure to avoid the bones or cartilage can introduce bone/cartilage chips into meat, introducing a foreign material into the consumer meat product. Today, this demanding task is performed by expert humans in grueling conditions --- each processing plant requires this precise cut be manually repeated approximately 300,000 chickens each day~\cite{4412581} while standing, while temperatures are held between 40 to 50$^\circ$F.

Despite decades of progress in agricultural automation~\cite{4412581}, dating back to the 1990s~\cite{daley1999modeling}, reliable and high-throughput automated chicken deboning has remained largely elusive due to three crucial and related gaps in the relevant literature: \textit{Precise 6-DoF cutting}: The chicken shoulder is comprised of the humerus and coracoid bones (roughly spherical; see Fig.~\ref{fig:overview}A) connected by tendons, muscle and skin. To cut through the spacing between the humerus and scapula, the knife must not only translate but also rotate to align with the 3D cutting plane that is perpendicular to the line connecting the centers of the two spheres. This requires motion planning and control in a full 6-DoF task space, whereas many prior studies on robot cutting have considered cutting only in two-dimensional spaces (see \cite{Xu-RSS-23, 10611596, 10246024, 9196623}). \textit{Dynamic adaptability in a partially observable environment}: Before the knife enters the shoulder region, the bones are concealed beneath soft tissue, preventing precise trajectory planning in advance. The robot must therefore dynamical adapt and determine its motion based on online force feedback in order to avoid undesirable bone contacts. In contrast, prior work on cutting single-material objects has often assumed open-loop trajectories without online adaptation (\cite{4412581, Heiden-RSS-21, long:hal-02463414, 8793880, 6265969}). \textit{Multi-material interaction}: Errors in estimating the shoulder joint spacing could result in accidental cutting through cartilage/bone, and existing approaches to robot cutting typically do not model the deformations and fracture of multiple interacting deformable materials~\cite{Xu-RSS-23}.



Our work makes both systems-level and algorithmic contributions to train and deploy a reactive cutting policy that can dynamically adapt a given nominal trajectory based on force feedback to achieve precise 6 DoF knife control, with a particular emphasis on chicken deboning (see Fig.~\ref{fig:overview}).

Our first contribution involves the design and implementation of \textbf{a customized open-source multi-material cutting simulator} that abstracts chicken deboning and supports policy learning. 
Our simulator offers a lower-fidelity approximation, capturing only the most-impactful features of the deboning process (e.g., relative cutting forces, coupling and fracture of heterogeneous materials), and is integrated with modern reinforcement learning (RL) libraries for efficient training. Our design strategically avoids a compute-intensive full-fidelity model that renders learning intractable.

While our custom simulator enables low-cost training, design, and evaluation, simulation results might not always translate to reality. At the same time, evaluating on real-world chickens is neither scalable nor repeatable. A variety of practical reasons prohibit careful and controlled experiments: chickens exhibit natural variability, 
the ground truth configuration of chicken shoulder bones cannot be known apriori, and their configuration changes during cutting. To enable rigorous study and repeatable experimentation in the real-world, our second contribution involves the design and implementation of \textbf{a simplified reusable physical testbed for multi-material cutting}, modeled after the chicken shoulder. In this model, the bones are represented as two Reusable Clay spheres with a fixed relative position, ensuring the existence of a valid cutting trajectory, while the surrounding muscle is approximated as a cuboid of Kinetic Sand embedding the two spheres. These two materials exhibit distinct hardness levels, causing the knife to experience different cutting forces when interacting with them. This set-up allows one to precisely control the bone position and orientation in real-world experiments while preserving the essential multi-material characteristics of the bird shoulder deboning task.

Our third contribution involves the design and training of \textbf{a dynamically adaptive 6 DoF cutting policy} via RL-based residual policy training. During training and evaluation, our residual policy adaptively adjusts the given nominal knife trajectory based on force feedback to avoid cutting into the two bone spheres whose ground truth positions are not accurately known. To facilitate successful sim-to-real transfer, we discretize the force signals and incorporate it into the policy’s observation space. We find that our residual cutting policy can successfully learn to cut through the spacing between the two bone spheres while avoiding collision with the spheres in simulation. When combined with domain randomization, we show that our adaptive cutting policy can be trained entirely in sim and deployed zero-shot both in our physical testbed and for real chicken deboning.

We conducted rigorous experiments in both our custom-built simulator and our simplified physical model. Further, to ensure that our contributions ultimately translate to the target setting, we also conducted experiments on real chicken shoulders using our residual policy training and sim2real transfer pipeline. Our results demonstrate the need for force feedback and suggest that our method significantly outperforms previous open-loop automation approaches~\cite{6265969} by up to 4x in terms of success rate and bone avoidance.

In summary, our primary contributions include:
\begin{itemize}
    \item The first customized open-source simulator for multi-material cutting that supports coupling and fracture of heterogeneous materials, and reinforcement learning, enabling rapid prototyping and efficient training. 
    \item The first real-world reusable multi-material cutting testbed modeled after the chicken shoulder, enabling controlled and repeatable experiments. 
    \item The first learning-based approach to 6 DoF multi-material cutting that transfers zero-shot to both our physical testbed and real chicken shoulders.
\end{itemize}


\section{RELATED WORKS}

\subsection{Deformable Cutting Simulator}

\hspace{-\parindent}While several robotic learning simulators support the deformation of soft materials (\cite{DBLP:conf/corl/LiangMHCMF18, 6386109, Genesis, lin2024ubsoft, li2023dexdeform, xian2023fluidlab, pmlr-v155-lin21a, Xiang_2020_SAPIEN}), most lack the capability to simulate fracture. Approaches to fracture modeling are typically divided into mesh-based and mesh-free methods. Mesh-based techniques like the Finite Element Method (FEM) require complex and costly re-meshing at the fracture interface, often restricting their application to simple planar cuts (\cite{Heiden-RSS-21, 4797517}). In contrast, mesh-free methods such as the Material Point Method (MPM) (\cite{10.1145/3197517.3201293, 10.1145/3340259, 10.1145/3306346.3322949}), Position-based Dynamics (PBD) (\cite{han20202d, 7912283, liu2021real}) and Smoothed Particle Hydrodynamics (SPH) (\cite{10.1145/2019406.2019411}) represent objects as particle aggregates, which more naturally accommodate arbitrary fracture surfaces.

However, a critical limitation of most mesh-free simulators for robotics is the absence of force feedback. Notable exceptions exist, but they retain significant constraints. For instance, DiSECt \cite{Heiden-RSS-21} computes interaction forces but is limited to planar cuts, while RoboNinja \cite{Xu-RSS-23} provides force readings for arbitrary cuts but is restricted to simulating simple bi-material objects consisting of one deformable material and another rigid material. To address these gaps, our developed simulator leverages MPM with the CPIC \cite{10.1145/3197517.3201293} method. This allows us to simulate arbitrary cutting trajectories for complex objects composed of multiple deformable materials, all while providing the crucial force feedback necessary for robotic learning tasks.

\subsection{Robotic Single-Material Cutting}

\hspace{-\parindent}Robotic cutting has been explored across a diverse range of applications, including food preparation tasks like processing vegetables (\cite{10611596, fang2024rh20t, grannen2023stabilize, 10113472, 10246024, 8793880}), fruits (\cite{9551558, 8967988, 9035011}), dough \cite{shi2023robocook}, and meat \cite{long:hal-02463414}. Its application extends to specialized industrial and medical domains, from hot-blade cutting of polystyrene \cite{Søndergaard2016} to precise surgical procedures (\cite{10160917, 9215039, 7989275}). To enhance robotic perception and control, researchers have developed sensorized tools, such as smart knives that provide force and depth feedback \cite{9905495}, and have utilized multi-modal haptic data to improve robustness \cite{9035073}.

However, a common thread in these prior works is the focus on cutting objects composed of a single, relatively uniform material, often executed through pre-defined open-loop trajectories or simple, repetitive planar cuts. In contrast, our research addresses the more complex challenge of cutting objects composed of multiple distinct deformable materials requiring complex spatial knife paths. This task necessitates real-time, multi-modal sensory feedback to dynamically adapt the cutting strategy to, for instance, precisely incise one material without damaging an adjacent one.

\subsection{Robotic Multi-Material Cutting}

\hspace{-\parindent}While research into robotic cutting of multi-material objects is limited, a few notable studies exist. Early work in this area tackled the challenging problem of chicken deboning by developing highly specialized, automated machinery (\cite{4412581, 6265969}). However, these systems were rigid in their design, lacking the versatility to adapt to natural variations in chicken size or bone positioning.

More recent studies have employed learning-based approaches with sensory feedback. For instance, one system successfully scooped the flesh from a grapefruit without piercing the peel by using a trained classifier and a closed-loop controller to manage the 1D cutting depth \cite{9196623}. Similarly, the RoboNinja framework demonstrated impressive adaptability in cutting multi-material fruits like avocados by using a state estimator and an adaptive policy to control a 2D cutting plane \cite{Xu-RSS-23}. Despite its success on objects with varied core geometries, its approach is fundamentally constrained to a 2D workspace.

A critical limitation of these recent advancements is that they address the problem assuming low-dimensional action spaces (1D or 2D). This is insufficient for complex, three-dimensional tasks like chicken deboning, which requires navigating intricate geometries and dynamically adapting the cutting tool's full 6D pose. Therefore, the challenge of versatile, high-degree-of-freedom robotic cutting for multi-material objects remains an open problem.



\section{Methodology}

\hspace{-\parindent}In this work, we utilize residual policy for trajectory adaptation and achieve successful sim2real transfer on applications such as the chicken deboning problem. We first implement a multi-material deformable cutting simulator that supports coupling and fracture of heterogeneous materials, as well as realistic modeling of cutting forces. Using this simulator, we construct a chicken shoulder cutting environment corresponding to our simplified real world model and integrate it with Stable-Baselines3, a reinforcement learning (RL) framework. Within this environment, we train a reactive residual cutting policy, where the policy adaptively adjusts the knife’s motion in response to force feedback. To promote robust sim-to-real transfer, we discretize force feedback and incorporate it into the policy’s observation space. Finally, we evaluate the trained residual cutting policy on a reusable real world multi-material cutting model.

\subsection{Multi-Material Cutting Simulator}
\label{method_simulator}

\hspace{-\parindent}We develop a multi-material cutting simulation environment that supports both the modeling of deformable materials and the coupling between them. In our scene, bones and meat are represented using an elastoplastic continuum model simulated with MLS-MPM \cite{10.1145/3197517.3201293}, governed by the von Mises yield criterion. MPM-based methods naturally capture large deformations and topological changes, making them well-suited for cutting tasks. To differentiate materials and maintain simulation stability, we assign the following distinct physical parameters: $\lambda_m$ = $27.78Pa$, $\mu_m$ = $41.67Pa$, $\rho_m$ = $10^3kg/m^3$, $\lambda_b$ = $222.22Pa$, $\mu_b$ = $333.33Pa$, $\rho_b$ = $2.819 \times 10^3kg/m^3$, where $\lambda_m$, $\mu_m$, $\lambda_b$, and $\mu_b$ are Lamé parameters for meat material and bone material, respectively, and $\rho_m$ and $\rho_b$ the density for meat material and bone material.

The knife is represented as a time-varying signed distance field (SDF) and approximated as a thin cuboid to accelerate computation. Contact between the deformable materials and the rigid knife is modeled by computing the cuboid surface normals and applying Coulomb friction. Material separation during cutting is handled directly by MLS-MPM, which inherently supports sharp and clean splitting of material points in soft media.

Our simulator is implemented in Python and Taichi \cite{10.1145/3355089.3356506}, and the environment inherits from the OpenAI Gym interface to ensure compatibility with Stable-Baselines3 \cite{stable-baselines3}. This allows reinforcement learning policies to be trained directly using algorithms provided by Stable-Baselines3. To ensure stable and efficient training, we further apply large force and velocity clipping, which prevents non-converging simulations and guarantees smooth RL rollouts.

\subsection{Residual Cutting Policy Learning}
\label{sec:residual_policy}


\hspace{-\parindent}The residual cutting policy is trained within our customized simulator. 
We let the knife follow a nominal trajectory determined based on estimated bone positions. At each simulation timestep, the residual policy adjusts the 6D (position and orientation) nominal trajectory. The policy’s observation space includes the current knife 6D pose, the next nominal pose (one timestep ahead), the measured 3D cutting force, and the previous 6D residual action. Based on these inputs, the policy outputs a 6D residual adjustment, which is added to the nominal trajectory to generate the executed action. To ensure stability and prevent unrealistic motions, the total residual action is clipped before being applied. To bridge the simulation and reality gap, especially the cutting force, we discretize each component of the 3D force and use the discretized 3D force in the residual policy observation.

\textbf{Observation Space}: At each timestep, the residual policy observes the following information from the environment: the current knife pose $\xi_t$; one step future knife pose $\hat{\xi}_{t+1}$, retrieved from the nominal trajectory; current knife force sensor reading $f_t$; and residual pose $e^\xi_{t}$.

\textbf{Action Space}: The residual policy produces a delta residual pose as the action $a_t$, which is added to the residual pose $e^\xi_{t}$ to produce the next residual pose: $e^\xi_{t+1} = e^\xi_{t} + \eta a_t$, where $\eta$ is an action scaling factor. $e^\xi_{t+1}$ will be added to the one-step future knife pose $\hat{\xi}_{t+1}$ to adjust the nominal trajectory: $\tilde{\xi}_{t+1} = \hat{\xi}_{t+1} + e^\xi_{t+1}$. And $\tilde{\xi}_{t+1}$ will be the next knife target pose. The initial value of the residual pose $e^\xi_0$ is $\mathbf{0}$.

\textbf{Nominal Trajectory}: Nominal trajectory is interpolated from several waypoints which are determined based on estimated bone positions (refer to \cite{6265969} for details on anatomical chicken shoulder joint  prediction). For example, previous studies utilizes a shoulder prediction model parameterized by visual bird features to obtain the waypoints~\cite{6265969}.

\textbf{Reward}: Since we already have a nominal trajectory to roughly command the motion of the knife, we design a reward function to guide the residual policy to learn to adjust the nominal trajectory to avoid cutting into the bones, while also avoiding unnecessary adjustment: $R = R_{bone} + R_{action}$. $R_{bone}$ is the bone avoidance reward. $R_{bone} = C - \alpha_1 b$. $C$ is a constant value, $\alpha_1$ is the scaling parameter and $b \in \{0, 1\}$ is a binary value. $b = 0$ when the knife cuts the bone at the current time step, while $b = 1$ when the knife does not cut the bone at the current time step. This reward term is crucial for eliciting desired behavior. $R_{action}$ is the action penalty, which regularizes the behavior of the residual policy. $R_{action} = \alpha_2||e^\xi_t||_2^2$, where $\alpha_2$ is a scaling parameter and $e^\xi_t$ is the residual pose.

\subsection{Reusable Real-World Model and Testbed}
\label{real-world model}

\begin{figure}[t]
    \centering
    \includegraphics[width=0.8\columnwidth]{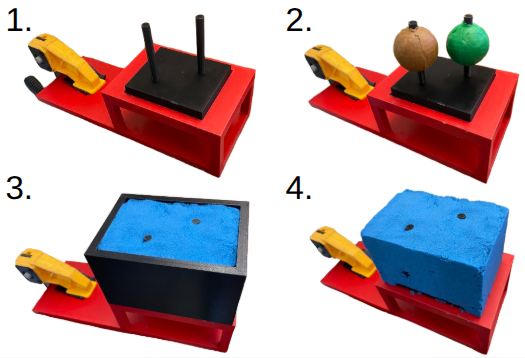}
    \caption{The setup process of our reusable physical testbed.}
    \label{fig:real_model_crafting}
\end{figure}

\hspace{-\parindent}We designed a simplified, reusable chicken shoulder model to enable quantitative and controllable study of the deboning problem. In this model, the bones are represented as two spheres with a fixed relative position, ensuring the existence of a valid cutting trajectory, while the surrounding soft tissue is approximated as a cuboid embedding the two spheres (Fig~\ref{fig:real_model_crafting}). 
We use Kinetic Sand to represent the soft tissue because it has stable physical properties \cite{Xu-RSS-23} and requires a relatively small force to be cut through. For the bone/cartilage material, which should allow fracture for simulating failure cases and require a larger cutting force to fracture than the soft tissue, we tested the cutting force of three different materials: Reusable Clay, Air-dry Clay and Play-Doh.
The average cutting force of the Re-usable Clay and the Air-dry clay are larger than that of Kinetic Sand, while the cutting force of Play-Doh is close to cutting Kinetic Sand.  
Therefore, Play-Doh is not a proper proxy for the bone material. We eventually chose Reusable Clay over Air-dry Clay for better efficiency, since Air-dry Clay requires over 24 hours of drying time and cannot be reused after fracturing.

We design and 3D-print a base and support for the deformable model (Fig. \ref{fig:real_model_crafting}). The base can be clamped to a table to maintain a fixed model position during cutting, while the support consists of two parallel cylindrical columns mounted on the base. Reusable clay spheres, crafted with a spherical mold, are placed on the columns to represent the shoulder bones with fixed relative positions.  The remaining volume is filled with Kinetic Sand using a box mold, representing the chicken muscle meat. 

\subsection{Sim2Real Transfer}
\label{sec:sim2real}

Due to the difficulty in resetting the real-world cutting environment and obtaining the environment states, we are unable to directly train RL policy in the real world. Instead, we utilize our custom-built simulator to train a policy and then deploy it to the real world. To zero-shot deploy the trained residual policy in the real world, we apply domain randomization. Specifically, we add noise into the proprioceptive observations of the residual policy while training in simulation. We add Gaussian noise $\mathcal{N}(0, 0.01)$ at the current knife position and add Gaussian noise $\mathcal{N}(0, 0.1)$ at the current knife orientation quaternion. We also observe that previous studies utilize discretization for successful sim2real transfer of tactile information \cite{qi2023general} on in-hand manipulation tasks. While our simulator captures the relative force trends necessary to distinguish between different materials during multi-material cutting, accurately modeling the absolute contact dynamics remains challenging, resulting in an inherent domain gap in the force profiles. To further bridge the gap between simulation force observation and real world force observation, we adopt discretization by normalizing and discretizing the force to make each orthogonal force component fall within the set $\{0, 0.1, 0.2, ..., 0.9, 1.0\}$. Crucially, our results demonstrate that by relying on this discretized force abstraction alongside proprioceptive domain randomization, a policy trained on simplified simulation dynamics can still successfully generalize to complex real-world environments.

\section{EXPERIMENTAL EVALUATION}
\label{sec:experiment}

\hspace{-\parindent} We conducted three experiments (one in simulation, one using our physical model, and one using real chicken shoulders) in order to answer the following questions:
\begin{itemize}
    \item \textbf{Q1}: How does our approach improve the cutting performance of nominal trajectory?
    \item \textbf{Q2}: How important is force feedback for adaptation?
    \item \textbf{Q3}: Can residual policy be successfully transferred to the real-world?
\end{itemize}

\textbf{Baselines}: We evaluated against two baselines that challenge our core designs.

\begin{figure}[h]
    \centering
    \includegraphics[width=3in]{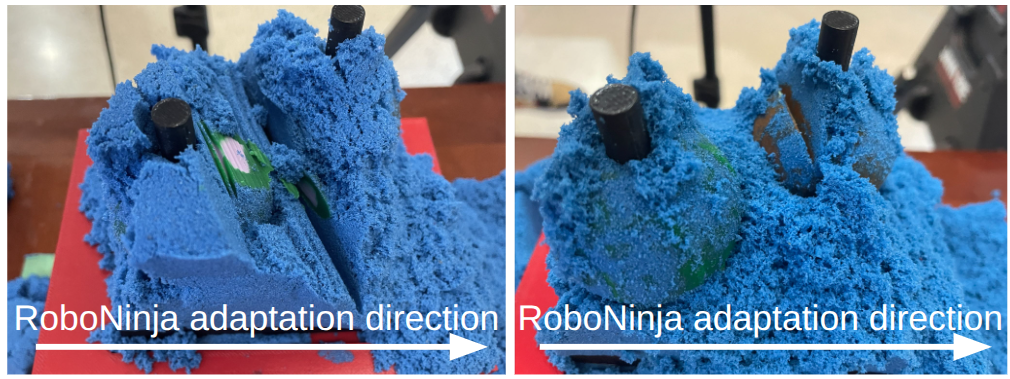}
    \caption{Roboninja's adaptive cutting policy qualitative evaluation results. The left picture shows that RoboNinja relies on knife-bone collision for interactive state estimation, which does crucial damage to non-rigid core in our task. The right is a failure case of RoboNinja when the knife first cuts into the right bone. Because RoboNinja can only adjust its trajectory to one direction (right), it keeps cutting into the right bone deeper and deeper.}
    \label{fig:roboninja}
\end{figure}

\begin{itemize}
    \item \textit{Nominal}: This baseline tracks the precomputed nominal trajectory to cut the deformable model.
    \item \textit{Adaptive w/o Force}: This baseline utilizes both nominal trajectory and residual policy. But the residual policy does not observe knife force. 
    \item \textit{Adaptive} (Ours): \textit{Adaptive} utilizes both nominal trajectory and residual policy mentioned in Section~\ref{sec:residual_policy}.
\end{itemize}
Note that RoboNinja~\cite{Xu-RSS-23} is not a suitable baseline for our approach given that it addresses a different and simpler problem.
Nevertheless, we experimented with extending  RoboNinja's adaptive cutting policy to our physical testbed. 
As expected, it could not complete our cutting task successfully due to three main reasons: (1) it relies on the detections of collisions to estimate the shape of the rigid core
but such collisions can severely damage non-rigid bones (see Fig.\ref{fig:roboninja}), (2) it can only adjust the trajectory to one direction, while our cutting task requires adjustment in arbitrary directions and orientation, and (3) it can produce large adjustments that miss the spacing between the two bone spheres. 

\textbf{Policy Learning}: We used the PPO \cite{schulman2017proximalpolicyoptimizationalgorithms} implementation from StableBaselines3 to train our residual policy. Both policy and value functions are parameterized via a MLP network with 2 hidden layers. The size of the hidden layers for both policy and value functions are (64, 64). The activation functions are all set as hyperbolic tangent. We use the same PPO hyper-parameters for all the baselines and our method ($\gamma$: 0.99, $\lambda$: 0.95, learning rate: 0.0003, clip range: 0.2, and batch size: 64). We train the polices on a computer with a single NVIDIA RTX 4090 GPU.

\textbf{Environment Setup:} As detailed in Sec.~\ref{method_simulator}, the physical simulation parameters remain identical when training the cutting policies for both the real-world proxy model and the actual chicken experiments.

\begin{figure}
    \centering
    \includegraphics[width=3in]{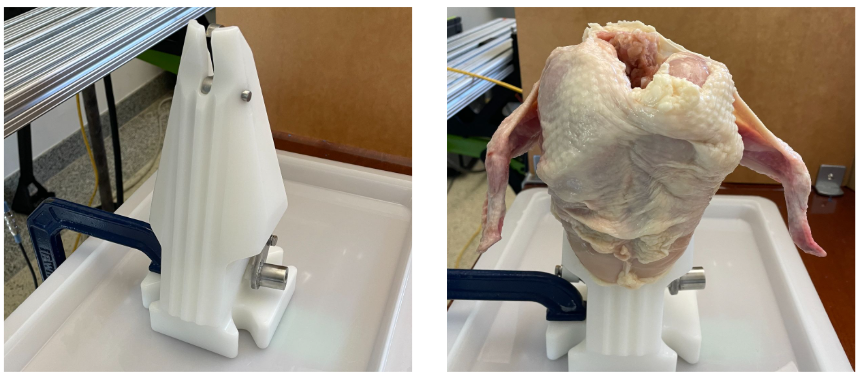}
    \caption{We use a cone which is clamped on the table to fix the chicken while conducting the cutting experiment.}
    \label{fig:chicken_exp_setup}
\end{figure}

In real-world model, the radius of the two bone spheres is $0.021m$ and the distance between the center of the two spheres is $0.05m$, hence leaving a $0.008m$ gap between the two sphere. To establish scenarios with varying estimation errors, we varied the 3D positions of the two spheres by $[\delta x, \delta y, \delta z]$, with sampling ranges $\delta x$: $[-0.005 m, 0.005 m]$, $\delta y$: $[-0.01 m, 0.01 m]$ and $\delta z$: $[-0.01 m, 0.01 m]$. The relative position of the two spheres remains constant, guaranteeing the existence of a valid cutting trajectory between the gap that avoids sphere (bone) contact. To maintain numerical stability, the simulated environment is scaled up by a factor of 5 relative to the physical real-world dimensions. For actual chicken experiment, the setup of chicken shoulder cutting experiment is shown in Fig.~\ref{fig:chicken_exp_setup}.

\textbf{Real-world Robot Setup}: Our real-world experiments are conducted on a UR5 robotic arm platform. A Robotiq FT 300 force–torque sensor is mounted between the robot wrist and a knife blade tool, providing force feedback during cutting. The sensor has a measurement range of 0–300~N, sufficient for capturing the interaction forces encountered in the multi-material cutting task.

\begin{figure*}
    \centering
    \includegraphics[width=0.9\textwidth]{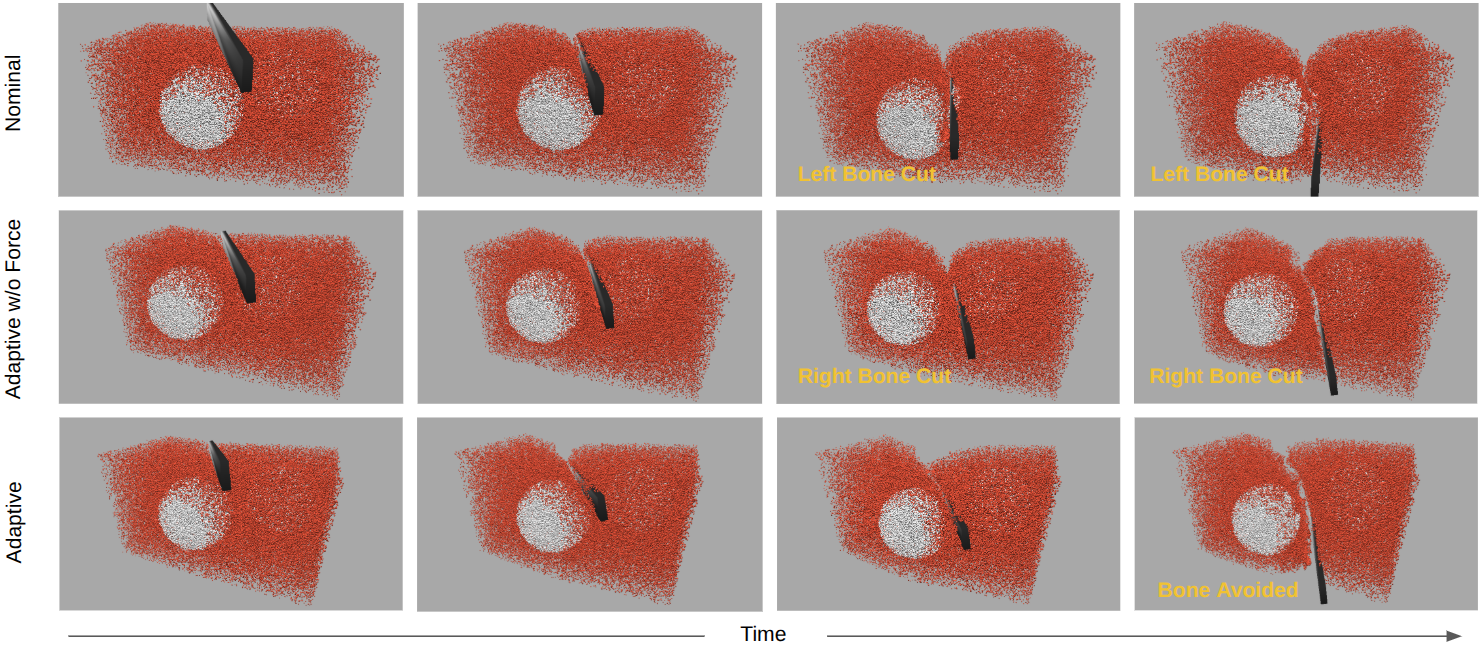}
    \caption{Simulation qualitative evaluation result. The first two rows show typical failure cases of the \textit{Nominal} and \textit{Adaptive w/o Force}. Due to lack of adaptivity, \textit{Nominal} rarely avoids the bone in most of the cases. Though \textit{Adaptive w/o Force} has the ability to adjust its trajectory, it can not perform well because it lacks necessary force feedback. The third row demonstrates a typical successful example of \textit{Adaptive}, which has both the ability of adaptation and receiving necessary force feedback. }
    \label{fig:sim_time_frame}
    \vspace{-3mm}
\end{figure*}

\textbf{Nominal Trajectory}: In simulation and physical model experiments, we generated the nominal trajectory by interpolating three waypoints. We held the three waypoints constant, and systematically varied the ground truth bone positions to reflect the fact that nominal trajectories are computed offline based solely on noisy estimates of bone positions.
For cutting real chicken shoulders, we used the nominal trajectory interpolated from waypoints (determined based on a shoulder prediction model parameterized by visual bird features \cite{6265969}).
We used cubic splines to interpolate the translation of the waypoints and spherical linear interpolation to interpolate between 3D rotations.

\subsection{Simulation Experiment}
\label{sim experiment}

We first evaluated each method on a set of 50 cuts in our simulator with randomly sampled bone positions.

\noindent \textbf{Metrics}: 
\begin{itemize}
    \item \textit{Violation Duration}: 
    To quantify bone avoidance, we count the number of timesteps during which the knife is cutting into the bones during each cut as the duration of violation. 
    \item \textit{Success Rate}: We consider an episode ``successful'' if there are no bone cuts in the episode. We consider it to be a bone cut if the knife is cutting into the bone during at least two consecutive environment steps. 
\end{itemize}


\noindent \textbf{Results}: We provide qualitative results in Fig.\ref{fig:sim_time_frame} and quantitative results in TABLE~\ref{table:sim_eval}. As shown, Open-loop \textit{Nominal} only results in 0.2 success rate and highest violation duration,  due to its lack of adaptivity, underscoring its sensitivity to partial observability and the need for adaptation in multi-material cutting tasks.
While \textit{Adaptive w/o Force} is able to perform better than \textit{Nominal} likely by improving the nominal trajectory while training in RL, it still lacks crucial force feedback. As expected, \textit{Adaptive} achieves the highest success rate and lowest bone cut count among all the baselines, demonstrating the effectiveness of residual policy training and the need for force-driven adaptivity.

\begin{table}[ht]
\caption{Simulation Evaluation Results }
\centering 
\resizebox{\columnwidth}{!}{
\begin{tabular}{c | c c} 
\hline 
Methods & Avg. Violation Duration (\%) ($\downarrow$) & Success Rate ($\uparrow$) \\ 
\hline 
\textit{Nominal} & $18.1 \pm 11.88$ & $0.20$ \\
\textit{Adaptive w/o Force} & $9.04 \pm 10.76$ & $0.54$ \\
\textit{Adaptive} & $\mathbf{4.74 \pm 9.66}$ & $\mathbf{0.80}$ \\
\hline 
\end{tabular}
} 
\label{table:sim_eval} 
\end{table}

\subsection{Real-world Model Experiment}
\label{real_world_simplified}

\begin{figure*}
    \centering
    \includegraphics[width=0.9\textwidth]{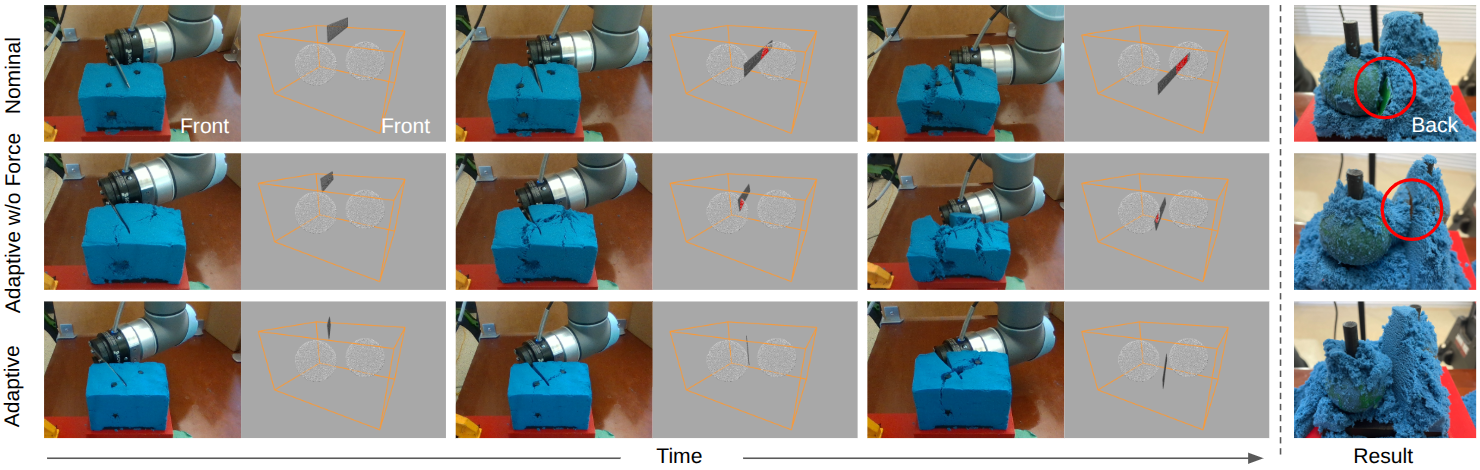}
    \caption{Real world model testbed qualitative evaluation result. To be able to ``see through'' the deformable model, we record robot end-effector trajectory and bone sphere position and visualize the cutting results using Taichi. The knife-bone intersection area will turn red in the visualization when a collision occurs. We also tear down the model after each cut to expose the bone and take a picture from the back. The pictures are shown in the last column. The red circle indicates the parts of the bone spheres that are cut off. }
    \label{fig:real_simplified_time_frame}
    \vspace{-3mm}
\end{figure*}



\hspace{-\parindent} We next evaluate each method over 20 cuts using our reusable physical model described in Section~\ref{real-world model} (see Section~\ref{sec:sim2real} our sim2real transfer approach).


\noindent \textbf{Metrics:}
\begin{itemize}
    \item \textit{Bone Cut Mass}: 
    The weight of reusable clay bone fragment(s) resulting from an unsuccessful cut, as measured by a jewelry scale with a resolution of 0.001 grams.
    \item \textit{Success Rate}: We consider a trial to be ``successful'' if no part of bones was cut during the trial. 
\end{itemize}

\begin{figure}
  \begin{minipage}[b]{.48\linewidth}
    \centering
    \includegraphics[width=0.9\linewidth]{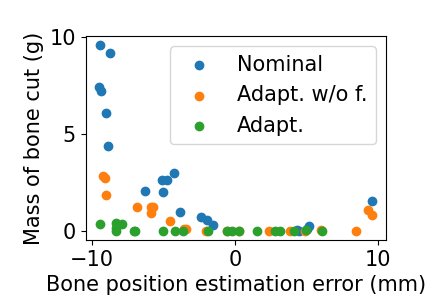}
    \captionof{figure}{\small{Bone cut mass a function of estimation error.}}
    \label{fig:estimation_error_vs_bone_mass}
  \end{minipage}\hfill
  \begin{minipage}[b]{.5\linewidth}
    \resizebox{\linewidth}{!}{
    \centering
    \begin{tabular}{c | c c}
    \hline
      Method & Avg. Bone & Success \\
       & Mass (g) & Rate \\
      \hline
      \textit{Nominal} & $3.032 \pm 3.090$ & $0.05$ \\
      \textit{Adapt. w/o f.} & $0.730 \pm 0.884$ & $0.35$ \\
      \textit{Adapt.} & $\mathbf{0.065 \pm 0.130}$ & $\mathbf{0.75}$ \\
      \hline
    \end{tabular}
    }
    \captionof{table}{\small{Results on Real-World Testbed.}}
    \label{table:simplified_eval}
  \end{minipage}
\end{figure}

\noindent \textbf{Results}: Please refer to Fig.\ref{fig:real_simplified_time_frame} for qualitative results, and TABLE~\ref{table:simplified_eval} for quantitative results. Similar to simulation results, our approach (\textit{Adaptive}) achieves the highest success rate and a substantially lower average bone cut mass than the two baselines. These results, not only reinforce the effectiveness of force-driven adaptation, but also demonstrate the impressive sim2real transfer performance of our learned residual policy. Further, taking advantage of our access to ground-truth bone position when using the simplified model, we studied the relation between the extent of estimation error and bone cut mass (see Fig.\ref{fig:estimation_error_vs_bone_mass}). Here, we approximate estimation error using the amount by which we shift the true bone position along the y axis (roughly perpendicular to the cutting direction), away from the estimated position. We observe that, in contrast to the two baselines struggle, \textit{Adaptive} still performs well even when the estimated bone positions are far from ground truth.

\begin{figure}[t]
    \centering
    \includegraphics[width=3in]{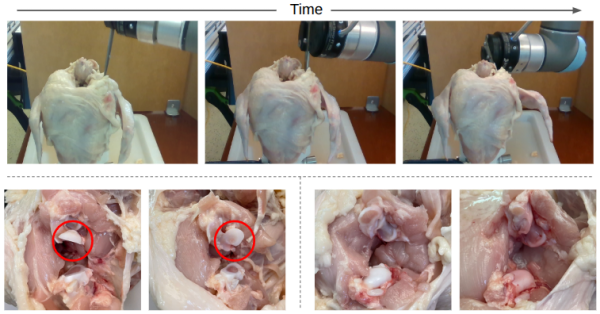}
    \caption{Top: Knife trajectory for cutting the chicken shoulder. Bottom left: typical cutting result of the \textit{Nominal} baseline, in which one large piece of the shoulder bone is cut off. Bottom right: example of utilizing the \textit{Adaptive}. The bone is successfully avoided and the knife precisely cuts between the two shoulder bones.}
    \label{fig:chicken_cutting}
    \vspace{-3mm}
\end{figure}

\subsection{Real Chicken Deboning Experiment}



\hspace{-\parindent}Finally, we evaluated each method on cutting real chicken shoulder joints. We first trained both learning-based methods in our custom-built simulator, 
while randomizing the position of the two simulated bone spheres according to the distribution of real chicken shoulder bone positions. We then used the sim2real transfer approach described in Section~\ref{sec:sim2real}.

\noindent \textbf{Metrics}: We use the same evaluation metrics as in Section \ref{real_world_simplified}. In addition, we also compute the ratio of cuts that bone fragments weighing under 0.05 grams since such small fragments are likely to be less dangerous.

\noindent \textbf{Baselines}: We use the nominal trajectory proposed in a previous robotic chicken deboning study~\cite{6265969} as a baseline. 

\noindent \textbf{Results}: We provide the qualitative and quantitative results of cutting real chicken shoulder in Fig.~\ref{fig:chicken_cutting} and TABLE.~\ref{table:chicken_exp}, respectively. Even on the challenging real-world chicken deboning task, our \textit{Adaptive} approach is able to considerably improve upon the nominal trajectory. These results also demonstrate our overall approach's ability to overcome dramatic sim2real gaps (simulated particles with computed forces to real chicken deboning with a force-torque sensor mounted on to a physical robot).

\begin{table}[ht]
\caption{Real Chicken Cutting Experiment Results}
\resizebox{\linewidth}{!}{
\centering 
\begin{tabular}{c | c c c } 
\hline 
Methods & Avg. Bone Mass (g) ($\downarrow$) & Success Rate ($\uparrow$) & Bone mass $<$ 0.05g ($\uparrow$) \\ 
\hline 
\textit{Nominal} & $0.339 \pm 0.316$ & $0.10$ & $0.20$ \\
\textit{Adaptive} & $\mathbf{0.121 \pm 0.239}$ & $\mathbf{0.50}$ & $\mathbf{0.70}$ \\
\hline 
\end{tabular}
}
\label{table:chicken_exp} 
\vspace{-3mm}
\end{table}

\section{CONCLUSIONS}

In this paper, we present the first multi-material deformable simulator that supports coupling and fracturing of heterogeneous materials and reinforcement learning, and the first real-world reusable multi-material cutting testbed modeled after the chicken shoulder. We use learning-based approach to tackle 6 DoF multi-material cutting tasks and achieve successful zero-shot transfer to both our physical testbed and real chicken shoulders. While our approach has demonstrated necessary adaptivity and transferability, it still has certain limitations for future extension. First, we have not tried our approach on more diverse bone shapes. Second, we rely on normalization and discretization of the force signal for sim2real transfer. We need to collect force data in advance and find the maximum value for manual normalization. Future research could consider cutting tasks with more arbitrary trajectory and more diverse bone shapes, especially more anatomically correct bone geometry. Other sim2real transfer techniques, such as Rapid Motor Adaptation can be applied to automatically adapt the trained policy to cutting tasks of different materials. Besides, we can improve the fidelity of our multi-material deformable simulator and even include more evaluation metrics for the multi-material cutting task.

\section{Acknowledgment}
This work is supported by the USDA National Institute of Food and Agriculture.

\addtolength{\textheight}{-0.0cm}   






\bibliographystyle{IEEEtran}
\bibliography{IEEEabrv,bibtex/main}

\end{document}